\documentclass[11pt,a4paper]{article}
\usepackage[hyperref]{acl2019}
\usepackage{times}
\usepackage{latexsym}
\usepackage{graphicx}
\usepackage{mathtools}
\usepackage{multirow}
\usepackage{tabularx}
\usepackage{arydshln}
\usepackage{color}
\usepackage{url}
\usepackage{tablefootnote}
\usepackage{bm}
\usepackage{booktabs}

\aclfinalcopy 

\title{\textsc{SemBleu}: A Robust Metric for AMR Parsing Evaluation}

\author{Linfeng Song \and Daniel Gildea \\
  Department of Computer Science\\ University of Rochester\\ Rochester, NY 14627
}

\date{}

\begin{document}
\maketitle
\begin{abstract}
Evaluating AMR parsing accuracy involves comparing pairs of AMR graphs.
The major evaluation metric, \textsc{Smatch} \cite{cai-knight:2013:Short}, searches for one-to-one mappings between the nodes of two AMRs with a greedy hill-climbing algorithm, which leads to search errors.
We propose \textsc{SemBleu}, a robust metric that extends \textsc{Bleu} \cite{papineni-EtAl:2002:ACL} to AMRs.
It does not suffer from search errors and considers non-local correspondences in addition to  local ones.
\textsc{SemBleu} is fully content-driven and punishes situations where a system's output does not preserve most information from the input.
Preliminary experiments on both sentence and corpus levels show that \textsc{SemBleu} has slightly higher consistency with human judgments than \textsc{Smatch}.
Our code is available at \url{http://github.com/freesunshine0316/sembleu}.
\end{abstract}

\section{Introduction}

\begin{figure}
    \centering
    \includegraphics[width=0.9\linewidth]{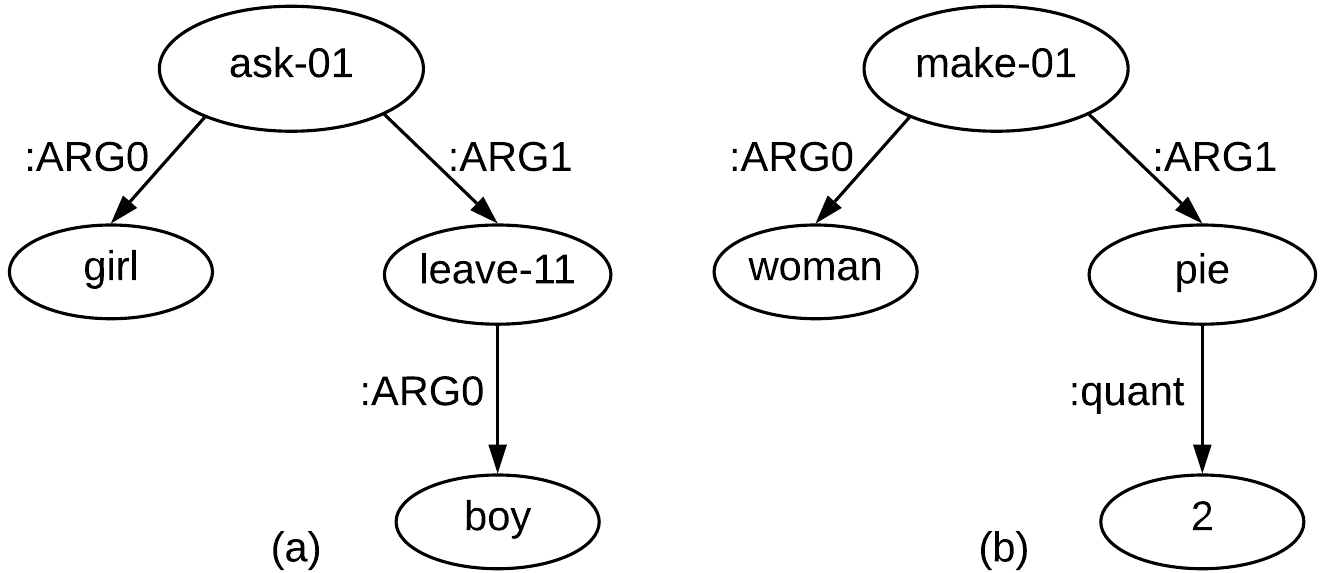}
    \caption{Two AMR examples meaning ``The girl asked the boy to leave.'' and ``The woman made two pies.'', respectively. Their \textsc{Smatch} score is 25 (\%).}
    \label{fig:example}
\end{figure}

Abstract Meaning Representation (AMR) \cite{banarescu-EtAl:2013:LAW7-ID} is a semantic formalism where the meaning of a sentence is encoded as a
rooted, directed graph.
Figure \ref{fig:example} shows two AMR graphs in which the nodes (such as ``girl'' and ``leave-11'') represent AMR concepts and the edges (such as ``ARG0'' and ``ARG1'') represent  relations between the concepts. 
The task of parsing sentences into AMRs has received increasing attention, due to the demand for better natural language understanding.

Despite the large amount of work on AMR parsing \cite{flanigan-EtAl:2014:P14-1,artzi-lee-zettlemoyer:2015:EMNLP,pust-EtAl:2015:EMNLP,K15-1004,buys-blunsom:2017:Long,konstas-EtAl:2017:Long,wang-xue:2017:EMNLP2017,ballesteros2017amr,P18-1037,peng:2018:ACL,P18-1170,D18-1198}, 
little attention has been paid to evaluating the parsing results, leaving \textsc{Smatch} \cite{cai-knight:2013:Short} as the only overall performance metric.
\citet{damonte-etal-2017-incremental} developed a suite of fine-grained performance measures based on the node mappings of \textsc{Smatch} (see below).

\begin{figure}
    \centering
    \includegraphics[width=0.9\linewidth]{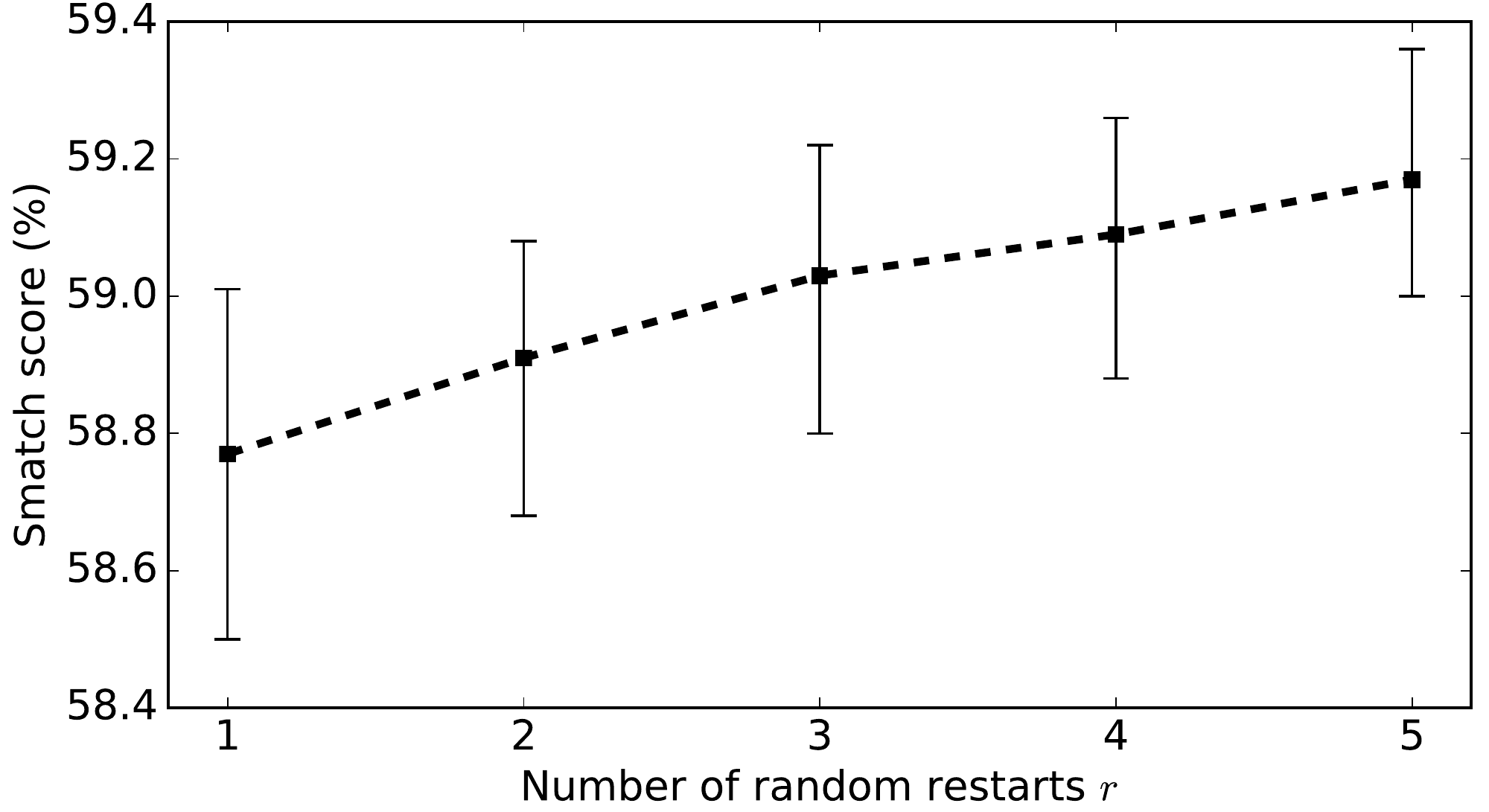}
    \caption{Average, minimal and maximal \textsc{Smatch} scores over 100 runs on 100 sentences. The running time increases from 6.6 seconds ($r$=1) to 21.0 ($r=4$).}
    \label{fig:errorbar}
    \vspace{-1.0em}
\end{figure}

\textsc{Smatch} suffers from two major drawbacks:
first, it is based on greedy hill-climbing to find a one-to-one node mapping between two AMRs (finding the exact best mapping is NP-complete).
The search errors weaken its robustness as a metric.
To enhance robustness, the hill-climbing search is executed multiple times with random restarts.
This decreases efficiency
and, more importantly, does not eliminate search errors.
Figure~\ref{fig:errorbar} shows the means and error bounds of \textsc{Smatch} scores as a function of the number of restarts $r$ over 100 runs on 100 sentences.
We can see that the variances are still significant when $r$ is large.
Furthermore, by corresponding with other researchers, we have learned that previous papers on AMR parsing report \textsc{Smatch} scores using differing values of $r$.

Another problem is that \textsc{Smatch} maps one node to another regardless of their actual content, and it only considers edge labels when comparing two edges.
As a result, two different edges, such as ``ask-01 :ARG1 leave-11'' and ``make-01 :ARG1 pie'' in Figure \ref{fig:example}, can be considered identical by \textsc{Smatch}.
This can lead to a overly large score for two completely different AMRs. 
As shown in Figure~\ref{fig:example}, \textsc{Smatch} gives a score of 25\% for the two AMRs meaning ``The girl asked the boy to leave'' and ``The woman made two pies'', which convey obviously different meanings.\footnote{\url{https://amr.isi.edu/eval/smatch/compare.html} gives more details.}
The situation could be worse for two different AMRs with few types of edge labels, where
the score could reach 50\% if all pairs of edges between them were accidentally matched.

To tackle the problems above, we introduce \textsc{SemBleu}, 
an accurate
metric for comparing AMR graphs.
\textsc{SemBleu} extends \textsc{Bleu} \cite{papineni-EtAl:2002:ACL}, which has been shown to be effective for evaluating a wide range of text generation tasks, such as machine translation and data-to-text generation.
In general, a \textsc{Bleu} score is a precision score calculated by comparing the $n$-grams ($n$ is up to 4) of a predicted sentence to those of a reference sentence.
To punish very short predictions, it is multiplied by a brevity penalty, which is less than 1.0 for a prediction shorter than its reference.
To adapt \textsc{Bleu} for comparing AMRs, we treat each AMR node (such as ``ask-01'') as a \emph{unigram}, and we take each pair of directly connected AMR nodes with their relation (such as 
``ask-01 :ARG0 girl'')
as a \emph{bigram}. 
Higher-order $n$-grams (such as 
``ask-01 :ARG1 leave-11 :ARG0 boy'')
are defined in a similar way.

\textsc{SemBleu} has several advantages over \textsc{Smatch}. 
First, it gives an exact score for each pair of AMRs without search errors.
Second, it is very efficient to calculate.
On a dataset of 1368 pairs of AMRs, \textsc{SemBleu} takes 0.5 seconds, while \textsc{Smatch} takes almost 2 minutes using the same machine.
Third, it captures high-order relations in addition to node-to-node 
and edge-to-edge 
correspondences.
This gives complementary judgments to the standard \textsc{Smatch} metric for evaluating AMR parsing quality.
Last, it does not give overly large credit to AMRs that represent completely different meanings.

Our initial evaluations suggest that \textsc{SemBleu} has higher consistency with human judgments than \textsc{Smatch} on both corpus-level and sentence-level evaluations.
We also show that the number of $n$-grams extracted by \textsc{SemBleu} is roughly linear in the AMR scale.
Evaluation on the outputs of several recent models show that \textsc{SemBleu} is mostly consistent with \textsc{Smatch} for results ranking, but with occasional disagreements.

\section{Our metric}

Our method is based on \textsc{Bleu}, which we briefly introduce, before showing how to extend it for matching AMR graphs.

\subsection{Preliminary knowledge on \textsc{Bleu}}

As shown in Equation \ref{eq:bleu_main}, the \textsc{Bleu} score for candidate $\boldsymbol{c}$ and reference $\boldsymbol{z}$ is calculated by multiplying a modified precision with a brevity penalty ($BP$).
\begin{equation} \label{eq:bleu_main}
    \textsc{Bleu}=BP \cdot \exp\left(\sum_{k=1}^{n} w_k \log p_k\right)
\end{equation}
$BP$ is defined as $e^{\textrm{min}\{1-\frac{|\boldsymbol{z}|}{|\boldsymbol{c}|}, 0\}}$, which gives a value of less than 1.0 when the candidate length ($|\boldsymbol{c}|$) is smaller than the reference length ($|\boldsymbol{z}|$). 
$p_k$ and $w_k$ are the precision and weight for matching $k$-grams, and $p_k$ is defined as
\begin{equation} \label{eq:bleu_prec}
p_k = \frac{|kgram(\boldsymbol{z}) \cap kgram(\boldsymbol{c})|}{|kgram(\boldsymbol{c})|} \textrm{,}
\end{equation}
where $kgram$ is the function for extracting all $k$-grams from its input.

\subsection{\textsc{SemBleu}}

To introduce \textsc{SemBleu}, we make the following changes to adapt \textsc{Bleu} to AMR graphs.
First, we define the size of each AMR ($G$) as the number of nodes plus the number of edges: $|G| = |G.nodes| + |G.edges|$.
This size is used to calculate the brevity penalty ($BP$).
Intuitively, edges carry important relational information.
Also, we observed many situations where a system-generated AMR preserves most of the concepts in the reference, but misses many edges.

Another change is to the $n$-gram extraction function ($kgram$ in Equation \ref{eq:bleu_prec}).
AMRs are directed acyclic graphs, thus we start extracting $n$-grams from the roots.
This is analogous to starting to extract plain $n$-grams from sentence left endpoints.
Note that the order of each $n$-gram is determined only by the number of nodes within it.
For instance, 
``ask-01 :ARG0 girl''
is considered as a bigram, \emph{not} a trigram.

Our $n$-gram extraction method 
adopts breadth-first traversal
to enumerate every possible starting node for extracting $n$-grams.
From each starting node $p$, 
it
extracts all possible $k$-grams ($1 \le k \le n$) beginning from $p$.
At each node, 
it
first stores the current $k$-gram 
before enumerating every descendant of the node and moving on.
Taking the AMR graphs in Figure \ref{fig:example} as examples, the $n$-grams extracted by our method are shown in Table \ref{tab:extracted}.

\begin{table}[t]
    \centering
    \begin{tabular}{ccl}
        Fg & $n$ & Extracted $n$-grams \\
        \midrule
\multirow{5}{*}{(a)} & 1 & ask-01; girl; leave-11; boy \\
        \cdashline{2-3}
                     & \multirow{3}{*}{2} & ask-01 :ARG0 girl; \\
                     &                    & ask-01 :ARG1 leave-11; \\
                     &                    & leave-11 :ARG0 boy; \\
        \cdashline{2-3}
                     & 3  & ask-01 :ARG1 leave-11 :ARG0 boy; \\
        \midrule
\multirow{5}{*}{(b)} & 1 & woman; make-01; pie; 2 \\
        \cdashline{2-3}
                     & \multirow{3}{*}{2} & make-01 :ARG0 woman; \\
                     &                    & make-01 :ARG1 pie; \\
                     &                    & pie :quant 2; \\
        \cdashline{2-3}
                     & 3  & make-01 :ARG1 pie :quant 2; \\
        \bottomrule
    \end{tabular}
    \caption{$n$-grams (separated by ``;'') extracted from the AMRs in Figure \ref{fig:example} with our extraction algorithm. \emph{Fg} represents the corresponding subfigure.}
    \label{tab:extracted}
\end{table}

\begin{table*}[t] \small
    \centering
    \begin{tabular}{rcccccc}
    Metric & CAMR vs JAMR & CAMR vs Gros & CAMR vs Lyu & JAMR vs Gros & JAMR vs Lyu & Gros vs Lyu \\
    \midrule
     \textsc{Smatch}  & 67.9 & 99.9 & 100.0 & 100.0 & 100.0 & 90.3 \\
    \textsc{SemBleu} & 69.0 & 99.9 & 100.0 & 100.0 & 100.0 & 90.9 \\
    \end{tabular}
    \caption{Corpus-level bootstrap accuracies (\%) for each system pair.}
    \label{tab:corpus_boot}
\end{table*}

\vspace{0.5em}
\textbf{Processing inverse relations}~~
One important characteristic of AMR is the inverse relations, such as ``ask-01 :ARG0 girl'' $\Rightarrow$ ``girl :ARG0-of ask-01'', for preserving the properties of being rooted and acyclic.
Both the original and inverse relations carry the same semantic meaning.
Following \textsc{Smatch}, we unify both types of relations by reverting all inverse relations to their original ones, before calculating \textsc{SemBleu} scores.

\vspace{0.5em}
\textbf{Efficiency}~~
As an important factor, the efficiency of \textsc{SemBleu}  largely depends on the number of extracted $n$-grams.
One potential problem is that there can be a large number of extracted $n$-grams for very dense graphs.
For a fully connected graph with $N$ nodes, there are $O(N^n)$ possible $n$-grams.
Luckily, AMRs are tree-like graphs \cite{chiang-cl18} that are very sparse.
For a tree with $N$ nodes, the number of $n$-grams is bounded by $O(n\cdot N)$, which is linear in the tree scale.
As tree-like graphs, we expect the number of $n$-grams extracted from AMRs to be roughly linear in 
the graph scale.
Our experiments empirically confirm this expectation.

\subsection{Comparison with \textsc{Smatch}}

In general, \textsc{Smatch} breaks down the problem of comparing two AMRs into comparing the smallest units: nodes and edges.
It treats each AMR as a bag of nodes and edges,
and then calculates an F1 score regarding the correctly mapped nodes and edges.
Given two AMRs, \textsc{Smatch} searches for one-to-one mappings between the graph nodes by maximizing the overall F1 score, and the edge-to-edge mappings are automatically determined by the node-to-node mappings.
Since obtaining the optimal mapping is NP-complete (by reduction from subgraph isomorphism), it uses a greedy hill-climbing algorithm to find a mapping, which is likely to be suboptimal.

One key difference is that \textsc{SemBleu} generally considers more global features than \textsc{Smatch}.
The only features that both metrics have in common are the node-to-node correspondences (also called unigrams for \textsc{SemBleu}).
Each bigram of \textsc{SemBleu} consists two AMR nodes and one edge that connects them, thus the bigrams already capture larger contexts than \textsc{Smatch}.
In addition, the higher-order $n$-grams of \textsc{SemBleu} capture even larger correspondences.
This can be a trade-off.
Generally, more high-order matches indicate better parsing performance, but sometimes we want to give partial credit for distinguishing partially correct results from the fully wrong ones.
As a result, combining \textsc{Smatch} with \textsc{SemBleu} may give more comprehensive judgment.

Another difference is the way to determine edge (relation) 
equivalence.
\textsc{Smatch} only checks edge labels, thus two edges with the same label but conveying different meanings can be considered as equivalent by \textsc{Smatch}.\footnote{One example is shown in the \textsc{Smatch} tutorial \url{https://amr.isi.edu/eval/smatch/tutorial.html}.}
On the other hand, \textsc{SemBleu} considers not only the edge labels but also the content of their heads and tails, as shown by the extracted $n$-grams in Table \ref{tab:extracted}.

Take the AMRs in Figure \ref{fig:example} as an example, \textsc{Smatch} maps ``girl'', ``ask-01'' and ``leave-11'' in (a) to ``woman'', ``make-01'' and ``pie'' in (b).
As a result, it considers that 
``ask-01 :ARG0 girl'' 
and 
``ask-01 :ARG1 leave-11'' 
in (a) are correctly mapped to 
``make-01 :ARG0 woman'' 
and 
``make-01 :ARG1 pie'' 
in (b), which does not make sense.
Conversely, \textsc{SemBleu} does not consider that these edges are correctly matched.

\section{Experiments}

We compare \textsc{SemBleu} with \textsc{Smatch} on the outputs of 4 systems over 100 sentences from the testset of LDC2015E86.
These systems are: \emph{CAMR},\footnote{\url{https://github.com/c-amr/camr}}
\emph{JAMR},\footnote{\url{https://github.com/jflanigan/jamr}} 
\emph{Gros} \citep{P18-1170} and \emph{Lyu} \citep{P18-1037}.
For each sentence, following \citet{callison2010findings}, annotators decide relative orders instead of a complete order over all systems.
In particular, 4 system outputs are randomly grouped into 2 pairs to make 2 comparisons.
For each pair, we ask 3 annotators to decide which one is better and choose the majority vote as the final judgment.
All the annotators have several years experience on AMR-related research, and the judgments are based on their impression on how well a system-generated AMR retains the meaning of the reference AMR.
Out of the 200 comparisons, annotators are fully agree on 142, accounting for 71\%.
With the judgments, we study consistencies of both metrics on sentence and corpus levels.

We consider all unigrams, bigrams and trigrams for \textsc{SemBleu}, and the weights ($w_k$s in Equation \ref{eq:bleu_main}) are equivalent (1/3 for each).
For sentence-level evaluation, we follow previous work to use NIST geometric smoothing \citep{chen2014systematic}. 
Following \textsc{Smatch}, inverse relations such as ``ARG0-of'', are reversed before extracting $n$-grams for making comparisons.

\subsection{Corpus-level experiment}

For corpus-level comparison, we assign each system a human score 
equal to the number of times that system's output was preferred.

Our four systems achieved human scores of 30, 33, 63 and 74.
They achieved \textsc{SemBleu} scores of 28, 30, 38 and 41, respectively,
and \textsc{Smatch} scores of 56, 56, 63 and 67, respectively.
\textsc{SemBleu} is generally more consistent with the human judgments.
In particular, there is a tie between \emph{CAMR} and \emph{JAMR} for \textsc{Smatch} scores, while \textsc{SemBleu} scores are more discriminating.
We use the script-default 2 significant digits when calculating \textsc{Smatch} scores, as their variance can be very large (Figure \ref{fig:errorbar}).
To make fair comparison, we also use 2 significant digits for \textsc{SemBleu}.

\vspace{0.5em}
\textbf{Bootstrap tests}~~
To conduct more comprehensive comparisons, we use bootstrap resampling \citep{koehn2004statistical} to obtain 1000 new datasets, each having 100 instances.
Every dataset contains the references, 4 system outputs and the corresponding human scores.
Using the new datasets, we check how frequently \textsc{SemBleu} and \textsc{Smatch} are consistent with human judgments on the corpus level as a way to perform significant test.

Table \ref{tab:corpus_boot} shows the accuracies of both metrics across all 6 system pairs (such as \emph{CAMR vs Lyu}).
Overall, \textsc{SemBleu} is equal to or slightly better than \textsc{Smatch} across all system pairs.
The advantages are not significant at $p<.05$, perhaps because of the small data size, yet human judgments on large-scale data is very time consuming.
Comparatively, the precisions of both metrics on \emph{CAMR vs JAMR} is lower than the other system pairs.
It is likely because the gaps of this system pair on both human and metric scores are much smaller than the other system pairs.
Still, \textsc{SemBleu} is better than \textsc{Smatch} on this system pair, showing that it may be more consistent with human evaluation.

\subsection{Sentence-level experiment}

\begin{table}
    \centering
    \begin{tabular}{rc}
    Metric  & Percent (\%) \\
    \midrule
    \textsc{Smatch} & 76.5 \\
    \textsc{SemBleu} & \textbf{81.5} \\
    \hdashline
    \textsc{SemBleu} ($n$=1) & 69.5 \\
    \textsc{SemBleu} ($n$=2) & 78.0 \\
    \textsc{SemBleu} ($n$=4) & 80.0 \\
    \bottomrule
    \end{tabular}
    \caption{Sentence-level accuracies, where the highest $n$-gram order is set to 3 by default, unless specified.}
    \label{tab:sentence}
\end{table}

For sentence-level comparison, we calculate the frequency with which a metric is consistent with human judgments on a pair of sentences.
Recall that we make two pairs out of the 4 outputs for each sentence, thus there are 200 pairs in total.

As shown in the upper group of Table \ref{tab:sentence}, \textsc{SemBleu} is 5.0 points better than \textsc{Smatch}, meaning that it makes 10 more correct evaluations than \textsc{Smatch} over the 200 instances.
This indicates that \textsc{SemBleu} is more consistent with human judges than \textsc{Smatch}.
The lower group shows \textsc{SemBleu} accuracies with different order $n$.
With only unigram features (node-to-node correspondences), \textsc{SemBleu} is much worse than \textsc{Smatch}.
When incorporating bigrams and trigrams, \textsc{SemBleu} gives consistently better numbers, demonstrating the usefulness of high-order features.
Further increasing $n$ leads to a decrease of accuracy.
This is likely because humans care more about the whole-graph quality than occasional high-order matches.

\subsection{Analysis on $n$-gram numbers}

\begin{figure}
    \centering
    \includegraphics[width=\linewidth]{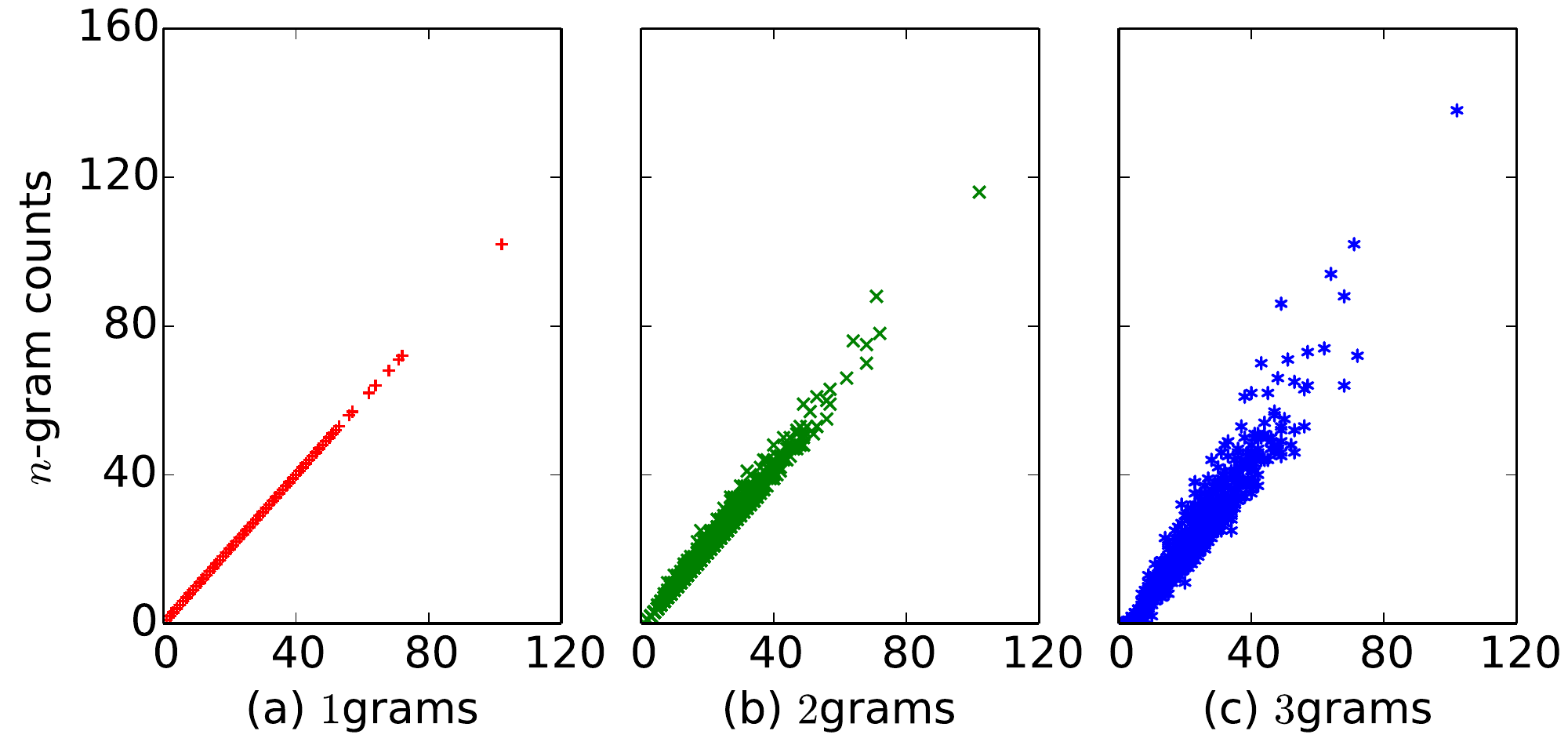}
    \caption{Extracted $n$-grams as a function of the number of AMR graph nodes.}
    \label{fig:ngrams}
\end{figure}

Figure \ref{fig:ngrams} shows the number of extracted $n$-grams
as a function of the number of AMR nodes on the devset of the LDC2015E86 dataset, which has 1368 instances.
The number of extracted unigrams is exactly the number of AMR nodes, which is expected.
The data points become less concentrated from bigrams to trigrams.
This is because the number of $n$-grams depends on not only the graph scale, but also how dense the graph is.
Overall, the amount of extracted $n$-grams is roughly linear in the number of nodes in the graph.

\subsection{Evaluating with \textsc{SemBleu}}

\begin{table} \small
    \centering
    \begin{tabular}{lrcc}
        Data & Model & \textsc{SemBleu} & \textsc{Smatch}\\
        \midrule
        \small{\multirow{5}{*}{LDC2015E86}} & Lyu  & 52.7 & \phantom{$\dagger$}73.7$\dagger$ \\
                             & Guo  & 50.1 & \phantom{$\dagger$}68.7$\dagger$ \\
                             & Gros & 50.0 & \phantom{$\dagger$}70.2$\dagger$ \\
                             & JAMR & 46.8 & 67.0 \\
                             & CAMR & 37.2 & 62.0 \\
        \midrule
        \small{\multirow{2}{*}{LDC2016E25}} & Lyu      & 54.3 & \phantom{$\dagger$}74.4$\dagger$ \\
                             & van Nood & 49.2 & \phantom{$\dagger$}71.0$\dagger$ \\
        \midrule
        \small{\multirow{4}{*}{LDC2017T10}} & Guo  & 52.0 & \phantom{$\dagger$}69.8$\dagger$ \\
                             & Gros & 50.7 & \phantom{$\dagger$}71.0$\dagger$ \\
                             & JAMR & 47.0 & 66.0 \\
                             & CAMR & 36.6 & 61.0 \\
        \bottomrule
    \end{tabular}
    \caption{\textsc{SemBleu} and \textsc{Smatch} scores for several recent models. $\dagger$ indicates previously reported result.}
    \label{tab:score_amrblue}
\end{table}

Table \ref{tab:score_amrblue} shows the \textsc{SemBleu} and \textsc{Smatch} scores several recent models.
In particular, we asked for the outputs of \emph{Lyu} \citep{P18-1037}, \emph{Gros} \citep{P18-1170}, \emph{van Nood} \citep{van2017neural} and \emph{Guo} \citep{D18-1198} to evaluate on our \textsc{SemBleu}.
For \emph{CAMR} and \emph{JAMR}, we obtain their outputs by running the released systems.
\textsc{SemBleu} is mostly consistent with \textsc{Smatch}, except for the order between \emph{Guo} and \emph{Gros}.
It is probably because \emph{Guo} has more high-order correspondences with the reference.

\section{Conclusion}

While one might expect a trade-off between speed and correlation with human
judgments, \textsc{SemBleu} appears to outperform \textsc{Smatch} in both dimensions.
The improvement in correlation with human judgments
comes from the fact that \textsc{SemBleu} considers larger fragments
of the input graphs.  The improvement in speed comes from avoiding the
search over mappings between the two graphs.  In practice, vertex mappings
can be identified by simply considering the vertex labels, and the labels
of their neighbors, through the $n$-grams in which they appear.
\textsc{SemBleu} can be potentially used to compare other types of graphs, including cyclic graphs.

\paragraph{Acknowledgments}

We are very grateful to Lisa Jin and Parker Riley for making annotations.
We thank Zhiguo Wang (Amazon AWS), Jinsong Su (Xiamen University) and the anonymous reviewers for their insightful comments.
Research supported by NSF award IIS-1813823.

\bibliography{acl2019}
\bibliographystyle{acl_natbib}

\end{document}